\documentclass{OAGMnotitle}
\OAGMarXiv{1304.1876}
\pdfoutput=1

\title{\"OAGM/AAPR 2013 -- The 37th Annual Workshop of the Austrian
  Association for Pattern Recognition}

\author{Justus Piater \and Antonio J. Rodr\'{i}guez S\'{a}nchez \\
  Institute of Computer Science, University of Innsbruck, Austria}

\begin{document}
\maketitle

The \href{https://oagm2013.iis.uibk.ac.at/}{37th Annual Workshop of
  the Austrian Association for Pattern Recognition} took place May
23--24, 2013, in the Plenary Hall of the new City Hall of Innsbruck,
Austria, under the motto \emph{Pattern Recognition and Computer Vision
  in Action}, and was attended by 56 participants.

The program was composed of 2 Keynote Talks by high-profile
researchers from outside of Austria, 7 Featured Talks by established
researchers and rising stars from the Austrian computer vision
community, 2 special talks by the two Microsoft Visual Computing Award
recipients, 11 contributed talks, and 9 posters.  The contributed
talks were selected from 14 submitted articles by peer review.  Each
article was reviewed by three members of the program committee. The 11
accepted articles are available on
\href{http://arxiv.org/abs/1304.1876}{arXiv}.

A Best Paper award was presented to Roland Perko for the paper
\begin{quotation}\noindent
  Roland Perko, Thomas Schnabel, Gerald Fritz, Alexander Almer, Lucas
  Paletta, \emph{Counting people from above: Airborne video based
    crowd analysis}.
\end{quotation}
A Best Student Paper award sponsored by the company M-BOX (Schwaz,
Austria) was presented to Tanja Schilling for the paper
\begin{quotation}\noindent
Tanja Schilling, Tom\'a\v{s} Pajdla, \emph{Euclidean Upgrade from a
  Minimal Number of Segments}.
\end{quotation}
The best papers were selected by the Conference Chairs based on the
reviewers' ratings and comments, as well as their own critical
jugdment.

The opportune moment was seized to announce and honor the two
recipients of the \href{http://aapr.icg.tugraz.at/award.php}{Microsoft
  Visual Computing Award} 2013:
\begin{quotation}
\noindent Chris Wojtan, IST Austria

\noindent Kristian Bredies, University of Graz
\end{quotation}

\vspace{3ex}
\noindent The \"OAGM/AAPR 2013 Workshop Chairs,\\
Justus Piater and Antonio J. Rodr\'{i}guez S\'{a}nchez

\clearpage
\section*{Keynote Speakers}

\begin{description}
\item[Jean Ponce] (INRIA and ENS, Paris, France): \textbf{Modeling
  visual recognition}

  \emph{Abstract --} This talk addresses the problem of automated
  visual recognition, that is, deciding whether an instance of some
  object class, for example, a chair, a person, or a car, is present
  in some picture, despite shape and color variations within the
  class, as well as viewpoint and illumination changes from one
  photograph to the next. After a brief historical discussion of the
  field, from early geometric methods to the current variations on
  bags of features and deformable part models, I will discuss some of
  our recent work that addresses issues such as taking explicitly into
  account shape and/or viewpoint variations within a category,
  selecting discriminative parts, and handling weak forms of
  supervision in tasks such as image categorization, object detection,
  image cosegmentation, and video interpretation.

  \emph{Biography --} Jean Ponce received the Doctorat de Troisieme
  Cycle and Doctorat d'Etat degrees in Computer Science from the
  University of Paris Orsay in 1983 and 1988. He has held Research
  Scientist positions at the Institut National de la Recherche en
  Informatique et Automatique, the MIT Artificial Intelligence
  Laboratory, and the Stanford University Robotics Laboratory, and
  served on the faculty of the Dept.\ of Computer Science at the
  University of Illinois at Urbana-Champaign from 1990 to 2005. Since
  2005, he has been a Professor at Ecole Normale Superieure in Paris,
  France, where he now also serves as Head of the Department of
  Computer Science. In 2003, Dr. Ponce was named an IEEE Fellow for
  his contributions to Computer Vision, and he received a US patent
  for the development of a robotic parts feeder. He has served on the
  editorial boards of Computer Vision and Image Understanding,
  Foundations and Trends in Computer Graphics and Vision, the IEEE
  Transactions on Robotics and Automation, the International Journal
  of Computer Vision (for which he served as Editor-in-Chief from 2003
  to 2008), and the SIAM Journal on Imaging Sciences. He was Program
  Chair of the 1997 IEEE Conference on Computer Vision and Pattern
  Recognition and served as General Chair of the year 2000 edition of
  this conference. He also served as General Chair of the 2008
  European Conference on Computer Vision. Dr.~Ponce is the co-author
  of Computer Vision: A Modern Approach, a textbook that has been
  translated in Chinese, Japanese, and Russian, and whose second
  edition came out in 2011.
\item[Fran\c{c}ois Fleuret] (IDIAP, Martigny, Switzerland): \emph{Boosting
  in Large Dimension Feature Spaces}

  \emph{Abstract --} It has been shown repeatedly that combining
  multiple types of features is an efficient strategy to improve the
  performance of machine learning techniques, particularly in computer
  vision. However, the use of multiple families of features increases
  the computational cost during training, which is usually linear with
  the feature space dimension.

  I will present in this talk different strategies to reduce that
  computational cost in the case of Boosting. This classical learning
  procedure builds a strong classifier iteratively by picking at every
  iteration a weak learner to minimizes a loss in a greedy
  manner. Dealing with feature spaces of large dimensions is usually
  achieved by sub-sampling a few features instead of using them
  all. The techniques we have developed use knowledge accumulated over
  previous Boosting iterations, or prior to starting the learning, to
  bias the sampling efficiently toward sub-families of features which
  are likely to provide a good reduction of the loss.

  Experiments on several data sets demonstrate that such strategies
  are far more aggressive during training and indeed reduce the loss
  more efficiently, often by one or two orders of magnitude.

  \emph{Biography --} Fran\c{c}ois Fleuret got the PhD degree in
  mathematics from the University of Paris VI in 2000 and the
  habilitation degree in applied mathematics from the University of
  Paris XIII in 2006. He has held positions at the University of
  Chicago, USA, and at the INRIA, France. Since 2007, he is the head
  of the Computer Vision and Learning group at the Idiap research
  institute in Switzerland, and is faculty member at the \'Ecole
  Polytechnique F\'ed\'erale de Lausanne (EPFL) as Ma\^itre
  d'Enseignement et de Recherche since 2011. His research is at the
  interface between statistical learning and algorithmic, with a
  strong bias toward computer vision. He is the author/co-author of 60
  reviewed journal and conference papers, serves as Associate Editor
  for the Transactions on Pattern Analysis and Machine Intelligence
  (TPAMI), and is or was expert for the Netherlands Organization for
  Scientific Research, the Austrian Research Fund, the Finish Research
  Council, and the French National Research Agency. He is the
  coordinator of the MASH European project.
\end{description}

\section*{Featured Speakers}

\begin{description}
\item[Margrit Gelautz] (TU Vienna): \textbf{Recent Advances in Local
  Stereo Matching}

  \emph{Abstract --} In this talk we address recent research in the
  field of local stereo matching, with a focus on adaptive support
  weight algorithms. In particular, we present a new stereo matching
  algorithm based on fast cost volume filtering that achieves
  high-quality disparity maps at real-time frame rates. Furthermore,
  we show an evaluation study that compares the performance of
  different adaptive support weight aggregation schemes. Finally,
  potential applications of the stereo-derived disparity maps for 3D
  film post-processing are discussed.
\item[Vladimir Kolmogorov] (IST Austria): \textbf{Discrete optimization
  algorithms in computer vision}

  \emph{Abstract --} A Basic \emph{Linear Programming relaxation} is a
  popular approach for discrete optimization in computer vision. I
  will describe recent results that completely characterize classes of
  optimization problems (from a certain family) for which this
  relaxation is exact. These classes include submodular functions,
  bisubmodular functions, and some new tractable classes. I will then
  talk about potential applications of bisubmodular functions. Time
  permitting, I will switch topics and talk about pattern-based CRFs
  for sequence data.
\item[Christoph Lampert] (IST Austria): \textbf{Visual Scene
  Understanding}

  \emph{Abstract --} It is one of the holy grails in computer vision
  research to build automatic systems that can understand images on a
  similar level as humans can, i.e.\ answer questions such as ``What
  objects are visible in this scene?'', ``How do they interact?'', and
  even ``What is going to happen next?''. It is only recently that
  computer vision has made significant progress in these directions,
  mainly driven by the development of new machine learning techniques
  and the availability of large amount of image data. Im my talk I
  will give a short overview of the state-of-the-art in the field, and
  highlight some of the work done in my group at IST Austria.
\item[Thomas Pock] (TU Graz): \textbf{Non-smooth Convex Optimization
  for Computer Vision}

  \emph{Abstract --} In this talk, we propose and analyze a flexible
  and efficient first-order primal-dual algorithm that is particularly
  suitable for solving large-scale non-smooth convex optimization
  problems. The algorithm comes along with guaranteed convergence
  rates, which are known to be optimal for first-order methods. A
  further advantage of the algorithm is, that it can be efficiently
  parallelized on graphics processing units, hence allows to solve
  some problems even in real-time. We will show applications to
  several computer vision problems ranging from optical flow
  estimation to 3D reconstruction.
\item[Peter Roth] (TU Graz): \textbf{Mahalanobis Metric Learning for
  Image Classification}

  \emph{Abstract --} Mahalanobis metric learning was recently of
  highly scientific interest in both, machine learning and computer
  vision. The main idea is to exploit the (discriminative) structure
  of the data and to explicitly learn a new metric providing much more
  meaningful distance measures. These can later be used for learning
  more effective classifiers or for direct k-nearest-neighbor
  matching. Thus, we first discuss metric learning from a very general
  point of view and discuss when it could be a beneficial tool for
  classification tasks. Then, we give an overview of (a) specific
  popular approaches which have been successfully applied for
  different applications and (b) of novel methods especially trying to
  reduce the computational effort during training. Finally, to
  demonstrate the benefits and to highlight the differences between
  different approaches, we discuss results for specific tasks such as
  face and person recognition.
\item[Robert Sablatnig] (TU Vienna): \textbf{Multispectral Image
  Acquisition for Manuscript Research}

  \emph{Abstract --} Manuscript analysis and reconstruction has long
  been solely the domain of philologists who had to cope with complex
  tasks without the aid of specialized tools. Technical scientists
  were only engaged in recording and conservation of valuable
  objects. In recent years, however, interdisciplinary work has
  constantly gained importance, concentrating not on a few special
  tasks only, like the development of OCR software, but comprising an
  increasing amount of relevant interdisciplinary fields like material
  analysis and document reconstruction. Within the framework of the
  Austrian Science Fund project ``Critical Edition of the New Sinaitic
  Glagolitic Euchology (Sacramentary) Fragments with the Aid of Modern
  Technologies'', philologists, image processing specialists and
  chemists are working together in an endeavor to analyze and edit
  three Old Church Slavonic parchment codices written in Glagolitic
  script.

  Within our framework, a multi-spectral representation of the page
  acquired is the basis for our subsequent analyses since this data
  representation holds a great potential for increasing the
  readability of historic texts, especially if the manuscripts are
  (partially) damaged and consequently hard to read. The readability
  enhancement is based on a combination of spatial and spectral
  information of the multivariate image data, a so called Multivariate
  Spatial Correlation. Additionally, Independent Component Analysis
  (ICA) and Principal Component Analysis (PCA) have been successfully
  applied for the separation and enhancement of diverse
  writings. Results show, that the readability my be enhanced by
  70\%. Furthermore we apply layout analysis and the structure of the
  character is characterised by its skeleton and dissected into trokes
  and nodes for an automatic formal classification of static, i.e.\ as
  they look like, graphetic attributes. The automatic character
  feature classification represents the starting point from where we
  will single out those features for computer processing that are able
  to mark a character distinctly in order to facilitate script
  reconstruction, automatic amendments of (incompletely preserved)
  letters, and OCR.
\item[Michael Zillich] (TU Vienna): \textbf{RGBD Vision for Robotics}

  \emph{Abstract --} The introduction of afforable RGBD sensors like
  the Microsoft Kinect or Asus Xtion has had a large impact especially
  on (indoor) robot vision, replacing stereo or more complex range
  sensing devices. Many traditionally hard 2D vision problems suddenly
  became a lot more tractable. In this talk I will give an overview of
  ongoing work in our group on RGBD-based scene segmentation, object
  reconstruction, recognition, classification and tracking, as well as
  their applications to typical robotics tasks such as finding and
  fetching items.
\end{description}

\section*{Program Committee}

\begin{itemize}
\setlength{\itemsep}{0pt}\setlength{\parsep}{0pt}\setlength{\parskip}{0pt}
\item Albert Rothenstein -- York University
\item Arjan Kuijper -- Fraunhofer IGD
\item Axel Pinz -- EMT
\item Branislav Micusik -- AIT
\item Christian Breiteneder -- IMS, TU Vienna
\item Christian Eitzinger -- Profactor
\item Christoph Lampert -- IST Austria
\item Csaba Beleznai -- AIT
\item Damien Teney -- University of Innsbruck
\item Daniel Leitner -- Univeristy of Vienna
\item Danijel Sko\v{c}aj -- VICOS
\item Dmitry Chetverikov -- MTA SZTAKI
\item El-hadi Zahzah -- University of La Rochelle
\item Frank Lenzen -- University of Heidelberg
\item Franz Leberl -- ICG, TU Graz
\item Friedrich Fraundorfer -- ETH Zurich
\item Georg Langs -- MIT
\item Gerhard Paar -- Joanneum Research
\item Gonzalo Pajares -- Universidad Complutense of Madrid
\item Hanchen Xiong -- University of Innsbruck
\item Harald Ganster -- Joanneum Research
\item Heinz Mayer -- Joanneum Research
\item Horst Bischof -- ICG
\item Horst Wildenauer -- TU Vienna
\item Johann Prankl -- TU Vienna
\item Josef Jansa -- IPF, TU Vienna
\item Josef Scharinger -- JKU Linz
\item Lech Szumilas -- Research Institute for Automation and Measurement
\item Lucas Paletta -- Joanneum Research
\item Marcus Hennecke -- EFKON
\item Margrit Gelautz -- TU Vienna
\item Markus Vincze -- ACIN, TU Vienna
\item Martina Uray -- Joanneum Research
\item Martin Kampel -- TU Vienna
\item Matthias R\"uther -- TU Graz
\item Michael Bleyer -- TU Vienna
\item Michael Zillich -- TU Vienna
\item Neil Bruce -- University of Manitoba
\item Peter Einramhof -- TU Vienna
\item Peter Roth -- TU Graz
\item Radim \v{S}\'ara -- CMP, Czech Technical University
\item Raimund Leitner -- CTR
\item Robert Sablatnig -- CVL, TU Vienna
\item Roman Pflugfelder -- AIT
\item Sven Olufs -- TU Vienna
\item Thomas Hoyoux -- University of Innsbruck
\item Thomas Pock -- TU Graz
\item Vaclav Hlavac -- CMP, Czech Technical University
\item Walter Kropatsch -- PRIP, TU Vienna
\item Wilhelm Burger -- FH Hagenberg
\item Xun	Shi -- York University
\item Yll	Haxhimusa -- TU Vienna
\end{itemize}
\end{document}